# Learning an Augmented RGB Representation with Cross-Modal Knowledge Distillation for Action Detection

Rui Dai[1,2], Srijan Das[3], François Bremond[1,2]
[1]Inria   [2]Université Côte d'Azur   [3]Stony Brook University
[1,2]{name.surname}@inria.fr   [3]{name.surname}@stonybrook.edu

## Abstract

*In video understanding, most cross-modal knowledge distillation (KD) methods are tailored for classification tasks, focusing on the discriminative representation of the trimmed videos. However, action detection requires not only categorizing actions, but also localizing them in untrimmed videos. Therefore, transferring knowledge pertaining to temporal relations is critical for this task which is missing in the previous cross-modal KD frameworks. To this end, we aim at learning an augmented RGB representation for action detection, taking advantage of additional modalities at training time through KD. We propose a KD framework consisting of two levels of distillation. On one hand, atomic-level distillation encourages the RGB student to learn the sub-representation of the actions from the teacher in a contrastive manner. On the other hand, sequence-level distillation encourages the student to learn the temporal knowledge from the teacher, which consists of transferring the Global Contextual Relations and the Action Boundary Saliency. The result is an Augmented-RGB stream that can achieve competitive performance as the two-stream network while using only RGB at inference time. Extensive experimental analysis shows that our proposed distillation framework is generic and outperforms other popular cross-modal distillation methods in action detection task.*

## 1. Introduction

Learning representation from untrimmed videos for action detection is a challenging vision task. Action detection aims at categorizing all the frames corresponding to every action occurring in an untrimmed video. The two main challenges for action detection are to tackle composite action patterns and fine-grained details [47]. These challenges are especially difficult in cases of real-world scenarios, where actions are densely distributed and overlapping each other [63]. To address these challenges, a typical setting, called two-stream network [49], consists in combining RGB with additional modalities like optical flow [41, 62], 3D poses [68, 11] to take into account the complementary nature of each modality. However, using such setting is contingent upon the availability of multiple modalities and of expensive processing resources. The cost of computing additional modalities could be prohibitive, especially for long untrimmed videos. These constraints limit the usage of multi-modal action detection methods for real-world applications.

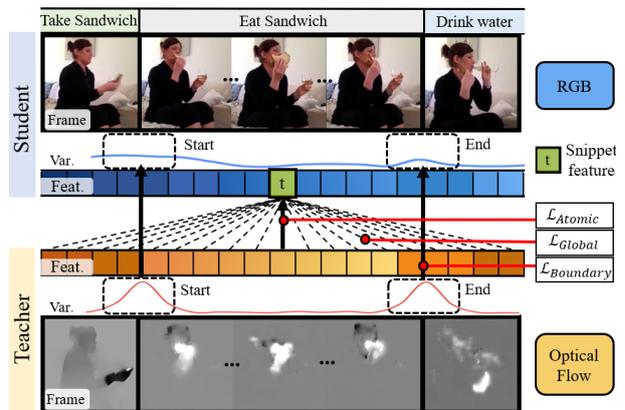

Figure 1. Proposed cross-modal distillation framework for action detection. Our distillation framework is composed of three loss terms corresponding to different types of knowledge to transfer across modalities. $\mathcal{L}_{Atomic}$: Atomic KD loss; $\mathcal{L}_{Global}$: Global Contextual Relation loss; $\mathcal{L}_{Boundary}$: Boundary Saliency loss.

Previous studies [20, 21] have shown that cross-modal Knowledge Distillation (KD) is an effective mechanism to avoid the computation of the additional modalities during test time, while preserving the complementary information from the additional modalities. However, most previous works [6, 18, 17] in the video understanding domain have investigated solely the classification of short trimmed videos [28, 50, 30]. In these works, each video corresponds to a single action and the distillation framework infuses the aggregated knowledge of an action instance from one modality into another. In contrast to trimmed videos, untrimmed videos contain rich sequential knowledge with complex temporal relations. Untrimmed videos in real-world scenarios tend to have cluttered background and mul-tiple correlated actions either in sequence [36] or in parallel [63, 48]. Therefore, distillation mechanisms tailored for classification tasks and extended for detection tasks lack in capturing fine-grained details along the temporal dimension. Now the question remains, what should be the right strategy to distillate cross-modal knowledge for action detection in untrimmed videos?

In this work, we propose a distillation framework to combine cross-modal information for detecting actions with high precision and minimal resource. The goal is to reach the two-stream performance while using only the RGB stream at inference time. The proposed distillation framework consists of a traditional teacher-student network architecture which operates in a Seq2Seq fashion [42, 10], thanks to three new distillation losses dedicated to the action detection task as illustrated in Fig. 1. The first loss in our formulation is the Atomic KD loss, which enables the RGB student network to mimic the feature representation of every individual snippet from the teacher network in a contrastive manner. This loss-term extends the cross-modal KD mechanism fabricated for the classification tasks to the temporal domain [38], by transferring the knowledge only between one-to-one corresponding snippets of different modalities. As a snippet is often shorter than the action instance in an untrimmed video, so this loss encourages a transfer of sub-representation [19] of the action, for example, "*raising arm*" in the "*drinking*" action. Here, such sub-representation w.r.t. the entire video corresponds to an atomic piece of knowledge within the complete action feature distribution. However, the untrimmed video is composed of a sequence of snippets, distilling only the atomic representation is not sufficient for learning discriminative action representation. Thus, distillation mechanisms dedicated to represent specifically an action within an untrimmed video are required.

We therefore introduce two loss-terms for sequence-level KD so as to transfer the cross-snippet relations between different modalities. Firstly, we propose a Global Contextual Relation loss to transfer the contextual information of the sequence between modalities. In our work, contextual information is defined as the embedding of the correlation between all the snippet features. Thanks to this loss term, every student snippet feature can learn in the latent space from all the correlated teacher snippets within the untrimmed videos (Fig. 1). With this loss-term, detecting one action in a snippet can benefit from the information in the correlated snippets (corresponding to related actions, e.g. *take and eat sandwich*) across modalities, resulting in better action detection performance. Secondly, we propose another KD loss to distillate the boundary saliency from the teacher to RGB student network, dubbed Boundary Saliency loss. This ensures a more precise action boundary detection of the RGB student which is prone to imprecise action boundary detection due to weak temporal signals. In an untrimmed video, the start and end moments of the action are more salient than other parts (see Fig. 1). Intuitively, the feature variation across consecutive snippets in the video can reflect such saliency of the action boundaries. Therefore, learning this variation from a modality that can better capture the movement (e.g. optical flow, 3D poses) encourages the RGB stream representation to be more sensitive to the action boundaries.

**Contributions.** To summarize, we take a step towards the little-explored but crucial, cross-modal KD for action detection. We build a Seq2Seq KD framework for action detection with a novel formulation. This formulation consists of an atomic-level KD loss and two sequence-level KD losses. The three loss terms in our formulation are jointly optimized in an end-to-end fashion. To the best of our knowledge, we are the first to propose a formulation containing sequential KD loss for the action detection task. We perform comprehensive experiments on five benchmarks. Our joint formulation significantly improves the vanilla RGB baseline (up to +6.8% improvement on MultiTHUMOS w.r.t. vanilla-RGB) and achieves the Two-stream performance while using only RGB at inference time. The consistent improvement on all these datasets corroborates the effectiveness and robustness of our distillation framework.

## 2. Related work

In this section, we briefly review the action detection architectures and the state-of-the-art cross-modal distillation methods.

### 2.1. Action Detection

Depending on the density of annotation, there are two kinds of dataset for action detection: (1) Sparsely labeled [26, 13, 36] and (2) Densely labeled [63, 48, 8] datasets. In the community, most action detection approaches [67, 16, 62] are custom-made for the sparsely labeled datasets. As densely labeled datasets include fine-grained actions occurring concurrently, they are more challenging and close to real-world scenarios [63]. Concerning the popular action detection approaches, anchor-based methods [61, 3, 16] are inspired by the two-stage object detection framework, which leverages a set of pre-defined anchors to generate action proposals along with another classification stage. However, anchor-based methods require a large number of anchors for generating the proposals and hence under-perform on densely labeled datasets [61, 5]. To handle the above issues, some methods [42, 31, 9, 51, 7] borrow the Seq2Seq framework from Natural Language Processing [12] to action detection. Seq2Seq methods are composed of a visual encoder to encode the primary spatio-temporal features, an efficient temporal filter to model the temporal information and a classifier to perform the frame-level action detection. This framework "interprets" the im-age sequence into a sequence of prediction scores. Frame-level action detection can be seen as a class-specific actionness detector. By referring to actionness-based methods [67, 60, 35], the action proposals can be generated from the frame-level detection results. Seq2Seq based approaches perform well on both types of dataset, especially for the densely annotated one [42, 10]. To build a generic distillation framework that can tackle both dataset types, we thus rely on the Seq2Seq paradigm. Furthermore to incorporate the appearance and motion based information, the state-of-the-art sequence based approaches make use of two-stream architectures [41, 42].

In order to avoid the expensive computation of two-stream architectures, cross-modal distillation methods [6, 18] and fast Optical Flow (OF) detectors [27, 52, 53, 66] have been introduced for video classification tasks. However, the fast OF detector techniques are modality-specific with relatively low performance [6]. Therefore, we propose to dig deeper into cross-modal distillation mechanisms for learning compact action representations for long continuous videos.

## 2.2. Cross-Modal Knowledge Distillation

The primary goal of Knowledge Distillation (KD) is to distill the information of a model learned from a teacher network into a student network. Many KD studies [23, 4, 40, 56, 37] explored transferring the knowledge from large complex models to small simpler models, i.e. model compression. In this work, we focus on cross-modal KD, where the difference between the teacher and student models mostly relies on input modalities rather than network architectures. In the video domain, Garcia et al. [18] developed a distillation framework for action classification with a four-step process that hallucinates depth features into RGB frames. Similarly, MARS [6] trains a RGB network in a single step, by back-propagating a linear combination of a OF distillation and classification losses through the entire network. Recently, Luo et al. [38] proposed a Graph Distillation (GD) method that can be applied to the action detection task. This method utilizes sliding windows to process untrimmed videos and distillates the knowledge of every window by minimizing the cosine distance in a mutual learning manner. GD aims at exploiting the privileged modalities and thus relies on a significant number of modalities. In contrast, our framework aims at effectively performing the distillation from the available modalities. Moreover, GD transfers knowledge only between the corresponding snippets (i.e. window), but does not consider the relations across snippets in the distillation, which is critical for handling a sequence of actions. Thus, to better tackle distillation for action detection, we propose a new formulation with three loss-terms. More specifically for transferring knowledge along the temporal dimension in the untrimmed video, we introduce a sequence-level distillation

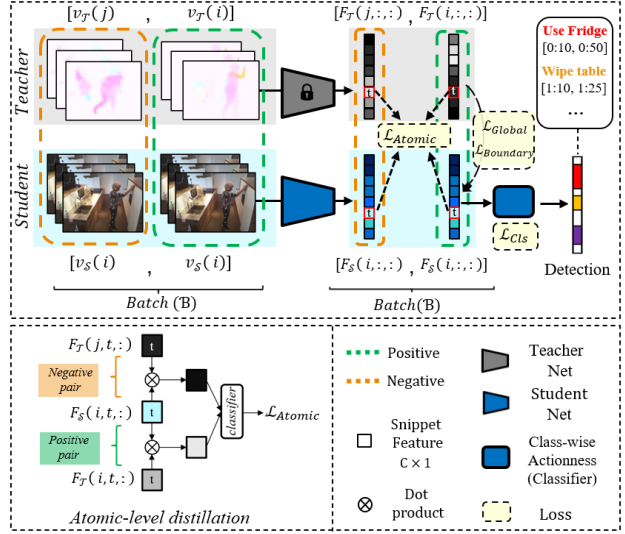

Figure 2. The proposed distillation framework. On the top, we present an example of a batch size ($\mathcal{B}$) of 2 untrimmed videos ($\mathcal{V}$) for both student ($\mathcal{S}$) and teacher ($\mathcal{T}$) networks. In this example, the input includes a pair of positive videos and a pair of negative videos. The sequence-level distillation and classification losses are employed only for positive pairs, while atomic-level distillation leverages both positive and negative pairs. On the bottom, we present the atomic-level distillation.

mechanism. Thanks to it, the network can be effective even with few additional modalities.

## 3. Proposed Distillation Framework

In this section, we first describe the overall architecture of our approach. We then detail the different losses in the proposed framework.

### 3.1. Overall Architecture

An overview of the architecture is shown in Fig. 2. In this work, the knowledge transfer occurs between the teacher and student networks. As discussed in Sec. 2.1, both networks are composed of a visual encoder and a temporal filter, following the Seq2Seq paradigm. For the visual encoder, we use I3D [2] to encode the spatio-temporal information of a snippet for RGB and Optical Flow (OF). Similar to previous action detection methods [14, 65], sequences of 16 frames are encoded to a single feature vector representation. The encoded feature maps of a video are then fed to the temporal filter. The choice of the temporal filter is flexible, since we can choose any well-known temporal model [42, 31, 24]. Here, we set a 5-layer SS-TCN [14] as default temporal filter, which is based on Dilated-TCN [31]. Both student and teacher have the same type of temporal filter with the same settings (i.e. dilation rate and channel size). In the training phase, the knowledge distillation is performed from the output feature of the teacher network towards the student network. Similar to [42, 10], this output feature map is further classified and grouped as a class-wise actionness detector for detecting the actions.

The input of teacher network is flexible to variant costly modalities (e.g. OF, 3D poses). By default, we chose the teacher network as OF stream, whereas the student network as RGB stream. In the following sections, we express the feature representation of a video indexed $i$ with $F_r(i, t, c)$, where $r \in \{\mathcal{T}, \mathcal{S}\}$ represents the teacher $\mathcal{T}$ and student $\mathcal{S}$; $t \in [1, T]$ represents the snippet index and $T$ the length of the video in snippets; $c \in \mathbb{Z}^C$ represents the channel index, $C$ is the channel size. This expression can be used for representing feature of a video or a snippet. For example, $F_r(i, :, :)$ and $F_r(i, t, :)$ represent the feature map of a video $i$ and the feature vector of a snippet for video $i$ at time step $t$, respectively. For an augmented RGB representation, the distillation is performed in two levels. First, we perform distillation at atomic-level to distillate the elementary representation of an action. Second, we perform a sequence-level distillation to distillate (i) the salient relations among the snippets, and (ii) the significant temporal variations across the snippets indicating action boundaries.

### 3.2. Atomic-level Distillation

To transfer the knowledge between two video sequence, firstly, we adapt and integrate the "representation loss" [38] in our overall formulation, dubbed Atomic KD loss. This loss term encourages the student to mimic the feature representation of every individual snippet feature of the teacher network. Our formulation is different from the previous work [38] that minimizes the cosine distance between the snippet features. Inspired by the recent success on contrastive learning [55, 22, 39], we build our model using a contrastive strategy to enhance the atomic-level knowledge imitation.

As shown in Fig. 2, let $[F_\mathcal{S}(i,t,:), F_\mathcal{T}(i,t,:)]$ represents a pair of training snippets from same video $i$ at time $t$ but across different modalities for the teacher and student networks. Let $F_\mathcal{T}(j,t,:)$ be another snippet representation from a randomly chosen video $j$ of the teacher stream and having a different label. We define the pair $[F_\mathcal{S}(i,t,:), F_\mathcal{T}(j,t,:)]$ as positive when $i = j$, otherwise negative. We aim at pushing closer the representations $F_\mathcal{S}(i,t,:)$ and $F_\mathcal{T}(i,t,:)$, while pushing apart $F_\mathcal{S}(i,t,:)$ and $F_\mathcal{T}(j,t,:)$, which can be seen as a binary classification task that tries to maximize the log-likelihood of the mutual information between the student and teacher representations. In practice, the loss is updated by batches with a batch size $\mathcal{B}$. If $\mathcal{N}$ negative pairs exist for each positive pair, then the number of samples in a batch of $\mathcal{P}$ positives is given by $\mathcal{B} = (\mathcal{N}+1)\mathcal{P}$ (see Fig. 2). To measure the mutual information between the student and the teacher, we compute:

$$\mathcal{L}_{Atomic} = \frac{1}{\mathcal{P}T} \sum_{i=j} \sum_{t=1}^{T} \log[\frac{\exp^{F_\mathcal{T}(j,t,:)^\top F_\mathcal{S}(i,t,:)}}{\exp^{F_\mathcal{T}(j,t,:)^\top F_\mathcal{S}(i,t,:)} + \phi}] + \frac{1}{T} \sum_{i \neq j} \sum_{t=1}^{T} [\log(1 - \frac{\exp^{F_\mathcal{T}(j,t,:)^\top F_\mathcal{S}(i,t,:)}}{\exp^{F_\mathcal{T}(j,t,:)^\top F_\mathcal{S}(i,t,:)} + \phi})] \quad (1)$$

where $\mathcal{P}T$ represents total number of positive snippets, $\phi$ is the ratio of the negative snippets to the cardinality of snippets in the training set. Note that, this loss term is accompanied by a linear combination with the other distillation losses and the class-wise entropy loss (i.e. supervised learning).

As the length of an action instance is often larger than a snippet, with atomic-level distillation, the teacher network transfers only the sub-representation of the actions [19]. Next, we propose a novel sequence-level distillation mechanism which has been neglected in the state-of-the-art methods.

### 3.3. Sequence-level Distillation

Sequence-level distillation transfers cross-snippet knowledge between different modalities in an untrimmed video by incorporating contextual information and taking benefit from the variations of cross-modal representation along action boundaries. Consequently, we propose two sequence-level distillation losses: (1) Global Contextual Relation, (2) Boundary Saliency, to improve action detection performance. Note that both sequence-level distillation losses are applied only between positive video pairs, corresponding to $\mathcal{P}$ videos.

#### 3.3.1 Global Contextual Relation

For sequence-level distillation, firstly, we propose to transfer contextual knowledge between modalities of the entire video. Intuitively, the detection of one given action could be supported by the detection of other related actions, which may be distant in the untrimmed video [41]. Hence, the representation of an action snippet could benefit from the contextual information across other snippets in the video pertaining to another modality. But the challenge in modeling such contextual relationships is the high complexity of the model for taking into account all the snippets in a video in relation with a single snippet. Therefore, we propose an embedding that projects the student-teacher features in a space where the global contextual relations among all actions are computed.

For the global contextual relation loss, we compute the Channel Covariance Matrix ($Cov$) of the sequence of snippets. Our motivation for using channel covariance is: (1) At each time instant, the channel contains the latent representation of actions, thus channel covariance embeds the relation between action events in a video; (2) Though the videos and action events have a high variation in length, the channel size is fixed. This results in channel covariance being a more robust representation.

Providing a feature map of the video, $Cov$ encodes the variance within each channel and the covariance between all channels over the whole video. Each element in the matrix reflects the correlation between two channels, which can characterize the specific activation patterns along time of an action class. Thus, the covariance matrix captures the relations between snippets along time and indicates whether a salient relation exists (i.e. which may be related to an action), while being computationally optimal. Here, the $Cov$ is formulated as:

$$Cov_r(i) = \frac{1}{T-1} \sum_{t=1}^{T} [F_r(i,t,:) - \mu_i][F_r(i,t,:) - \mu_i]^T \quad (2)$$

such that $r \in \{\mathcal{T}, \mathcal{S}\}$, and $\mu_i$ represents the mean value of all the channels in the feature map $F_r(i,:,:)$ of a video $i$. The covariance matrix $Cov_r \in \mathbb{R}^{C \times C}$ is a symmetric matrix and thus it is determined by $\frac{C(C+1)}{2}$ values. We apply a filter mask extracting all the entries on and above the diagonal of the covariance matrix. We reshape these values in the form of a vector $G_r(i)$:

$$G_r(i) = mask[Cov_r(i)] \quad (3)$$

where $mask(.)$ is the filter mask operation. The obtained feature vector $G_r(i)$ represents the channel covariance of the video. We then enforce a distillation loss in the embedded space from the frozen teacher to the student over the positive video pairs ($\mathcal{P}$). This is performed by minimizing the mean square error, which is formulated as the Global Contextual Relation loss:

$$\mathcal{L}_{Global} = \frac{1}{\mathcal{P}} \sum_{i=1}^{\mathcal{P}} ||G_\mathcal{T}(i) - G_\mathcal{S}(i)||^2 \quad (4)$$

The differential property of equation 2 enables to train our teacher-student framework jointly with the other losses.

### 3.3.2 Boundary Saliency

The boundary saliency loss term is used in our formulation to learn comparatively precise boundaries for action detection. In an untrimmed video, we find that the starting and ending of the action are more salient than other parts [34], that brings us crucial information to detect the transition of an action to another action or background. Intuitively, the sharp variation across consecutive snippets in the video can reflect such saliency of the action boundaries, which is a cross-snippet knowledge. Transferring the knowledge of feature evolution along time encourages the features to be more sensitive at the action start and end, thus assisting the class-wise actionness detector in the student network to detect precise boundaries of the action instances. Such an approach is especially effective when the modality processed at the teacher network provides pertinent boundary information. For instance, modalities which are sensitive to motion (e.g. OF, 3D poses) are able to bring a significant benefit from this loss term. In addition, this loss-term also encourages to retain the temporal consistency across the different modalities.

In practice, we first define the variation between consecutive snippets as $Var(i)$ for video $i$, which is formulated as:

$$Var_r(i) = \frac{1}{T-1} \sum_{t=1}^{T-1} \sum_{c=1}^{C} [F_r(i,t+1,c) - F_r(i,t,c)] \quad (5)$$

where $r \in \{\mathcal{T}, \mathcal{S}\}$. Then, we define the Boundary Saliency loss as the $L1$ distance between the frozen teacher and the student network over the $\mathcal{P}$ positive pairs, which is formulated as:

$$\mathcal{L}_{Boundary} = \frac{1}{\mathcal{P}} \sum_{i=1}^{\mathcal{P}} |Var_\mathcal{T}(i) - Var_\mathcal{S}(i)| \quad (6)$$

With both sequence-level distillation losses, the student network learns two types of cross-snippet information from the other modalities. Below, we summarize the training procedure section.

### 3.4. Training and Testing

To sum up, firstly, we train the teacher networks with the classification ($Cls$) loss, i.e cross-entropy. The weights of the teacher network is then frozen followed by training the student network. During training, multiple distillation losses are jointly optimized with classification loss for the end task, i.e. action detection. On one hand, the atomic distillation is trained in a contrastive manner (with positive and negative pairs), whereas the sequence-level distillation losses are performed in a non-contrastive manner by utilizing only the positive pairs in a batch. The overall objective is formulated as:

$$\mathcal{L}_{total} = \mathcal{L}_{Cls} + \alpha_1 \mathcal{L}_{Atomic} + \alpha_2 \mathcal{L}_{Global} + \alpha_3 \mathcal{L}_{Boundary} \quad (7)$$

where $\alpha_i$ are the loss weighting factors determined during the validation step. $\mathcal{L}_{Cls}$ represents the cross-entropy classification loss. We call the educated-student network as **Augmented-RGB**. During inference time, we only use RGB videos as input to detect the actions and up-sample the predicted logits to the same temporal resolution as the ground truth to perform the evaluation.

## 4. Experimental Analysis

To corroborate the effectiveness of our proposed KD framework, we perform an exhaustive experimental analysis for the action detection task.

### 4.1. Dataset Description

We evaluate our framework on five action detection datasets: Charades [47], PKU-MMD [36], TSU [8], THUMOS14 [26], and MultiTHUMOS [63]. These datasets contain videos of different types: (1) sport and daily living videos, (2) short and long videos, (3) densely and sparsely labeled videos. Note: there are two settings on Charades: (1) video-level action classification, (2) frame-level action detection (Charades_v1_localize [47]). We only target the second one in this paper.

Table 1. Ablation study for the proposed framework on Charades and PKU-MMD (CS) datasets. For PKU-MMD we consider IoU=0.1.

|  | $\mathcal{L}_{Atomic}$ | $\mathcal{L}_{Global}$ | $\mathcal{L}_{Boundary}$ | Charades | PKU-MMD |
|---|---|---|---|---|---|
| Teacher-OF | – | – | – | 18.6 | 68.4 |
| Vanilla-RGB | – | – | – | 22.3 | 79.6 |
| Two-stream | – | – | – | 24.8 | 83.4 |
| Atomic | ✓ | – | – | 23.9 | 82.7 |
| Sequence | – | ✓ | – | 23.8 | 83.7 |
|  | – | – | ✓ | 23.4 | 83.1 |
|  | – | ✓ | ✓ | 24.2 | 84.2 |
| Mixture | ✓ | ✓ | – | 24.4 | 84.3 |
|  | ✓ | – | ✓ | 24.2 | 83.7 |
| Total | ✓ | ✓ | ✓ | **24.6** | **85.5** |

All the datasets are evaluated by the mean Average Precision (mAP). We evaluate the per-frame mAP on densely labeled datasets following [63, 47].

### 4.2. Implementation Details

For extracting the additional modalities, Optical Flow (OF) is obtained using TVL1 [54], the 3D Poses are extracted using LCRNet++ [44]. In this work, we adapt the 5-layer SSTCN [14] as the temporal filter, the output channel size $C$ is 256. While training the teacher-student framework, we use Adam optimizer [29] with an initial learning rate of 0.001, and we scale it by a factor of 0.3 with a patience of 10 epochs. The network is trained for 300 epochs with a mini-batch $\mathcal{B}$ of 16 videos for Charades, 8 videos for PKU-MMD, THUMOS, and 4 videos for the TSU dataset. $\mathcal{N}$ is set to 1, $\mathcal{P}$ as $\frac{\mathcal{B}}{2}$ and $\alpha_i$=[300, 100, 5]. We use binary cross-entropy for multi-label classification (i.e. class-wise actionness). For sparsely-labeled datasets: THUMOS14 and PKU-MMD, following [38, 10], a post-processing step is performed to generate the action boundaries.

### 4.3. Ablation Study

Firstly, we discuss about the effectiveness of the losses proposed in our distillation framework. Tab. 1 shows the comparison of action detection performance on Charades and PKU-MMD (IoU=0.1). This table also shows the impact of progressively integrating the KD losses in our distillation framework. The vanilla-RGB is the network trained using only $\mathcal{L}_{Cls}$ without distillation. Compared to vanilla RGB, while training with $\mathcal{L}_{Atomic}$, $\mathcal{L}_{Global}$, $\mathcal{L}_{Boundary}$ independently obtains an improvement of +3.1, 4.1, 3.5% mAP on PKU-MMD respectively. The action detection performance is further improved by the convex combination of any two losses w.r.t. their individual counter-parts. This shows the complementary functionalities of the proposed losses. Also, note that the combination of the sequence losses contributes higher than the atomic loss. This observation supports the importance of the sequence-level losses for action detection. Finally, when trained with all the three losses, the student outperforms all the baselines (+2.3%, +5.9% w.r.t. vanilla RGB stream on Charades and PKU-MMD). These results show that both our design choices and different losses contribute to the overall performance of our approach.

In Tab. 2, we show that our distillation mechanisms perform better at feature-level than at logit-level. The primary reason behind this trend is that we are performing cross-modal distillation, where the frozen teacher may underperform compared to the student network (e.g. OF on Charades and PKU-MMD) w.r.t. the different modalities. As the logits represent the classification scores, they may introduce noise from the weak teacher via KD into the RGB student.

### 4.4. Analysis of our Distillation Framework

In this section, we further analyze our distillation framework in different aspects.

**Comparison with popular cross-modal KD methods:** Tab. 3 presents a comparison of our extended atomic distillation with state-of-the-art cross-modal KD methods, learning from OF. These baseline methods [25, 6, 18, 58] using traditional losses like MSE and cosine distance are actually designed for classification tasks. For the comparative analysis with our $\mathcal{L}_{Atomic}$, we adapt them following [38] for the task of action detection. $\mathcal{L}_{Atomic}$ consistently outperforms all the baseline methods on Charades and PKU-MMD datasets (+1.6%, +3.1% w.r.t. vanilla RGB stream on Charades and PKU-MMD).

**Performance with different Temporal filters:** In Tab. 4, our distillation framework is implemented with different temporal filters to confirm its robustness. The experiments are performed with a student network learning from a teacher network pre-trained with OF on Charades dataset. The results show that both SSTCN [14] and TGM [42] consistently improve the performance of RGB stream and achieve the competitive performance of two-stream network.

**Analyzing our framework with different modalities:** In Tab. 5, we validate that our proposed method is generic and can be effective with different modalities. For experimentation, we perform distillation from OF and 3D Poses. For 3D poses, the teacher consists of 2s-AGCN [46] as visual encoder followed by the temporal filters for detecting actions. In datasets like Charades, most actions involve human-object interactions with prominent motion patterns and in datasets like PKU-MMD, most actions have similar appearance with variant motion over time. Thus, OF stream provides more salient information than Pose stream on these datasets. Whereas 3D Poses are robust to the change of the view-points and thus, significantly improves the action detection performance in cross-view settings (see Tab. 5). Furthermore, with a multi-teacher network with OF and

Table 2. Feature-level and logit-level distillation. The student learns from OF stream. For PKU-MMD, we set IoU=0.1.

|  | Charades | PKU-MMD |
|---|---|---|
| Logit | 23.7 | 84.9 |
| Logit+Feature | 24.2 | 85.4 |
| Feature (Ours) | **24.6** | **85.5** |

Table 3. Comparison with cross-modal KD methods and $\mathcal{L}_{Atomic}$ on Charades and PKU-MMD datasets. For PKU-MMD, IoU=0.1.

|  | Charades | PKU-MMD |
|---|---|---|
| Vanilla-RGB | 22.3 | 79.6 |
| +$\mathcal{L}_{Hall}$ [18] | 22.7 | 81.5 |
| +$\mathcal{L}_{MARS}$ [6] | 23.5 | 81.7 |
| +$\mathcal{L}_{GD}$ [38] | 23.3 | 82.2 |
| +$\mathcal{L}_{Atomic}$ (Ours) | **23.9** | **82.7** |

Table 4. Ablation for different temporal filters: SS-TCN and TGM on the Charades. The models learn from the OF.

|  | SSTCN | TGM |
|---|---|---|
| Vanilla-RGB | 22.3 | 18.9 |
| Two Stream | 24.8 | 21.5 |
| Augmented-RGB | 24.6 | 21.2 |

Table 5. Ablation for different modalities on Charades, PKU-MMD (CS), TSU-CS and TSU-CV. For TSU, the reported values are frame-based mAP (%). The IoU threshold for PKU-MMD is 0.1.

|  | Charades | PKU-MMD | TSU-CS | TSU-CV |
|---|---|---|---|---|
| Teacher-OF | 18.6 | 68.4 | 29.4 | 17.5 |
| Teacher-Pose | 9.8 | 65.0 | 26.2 | 22.4 |
| Vanilla-RGB | 22.3 | 79.6 | 29.2 | 18.9 |
| Two-stream RGB + Pose | 23.0 | 82.9 | 32.6 | 23.7 |
| Two-stream RGB + OF | 24.8 | 83.4 | 33.5 | 19.5 |
| Pose Augmented RGB | 23.2 | 84.7 | 32.4 | 23.6 |
| OF Augmented RGB | 24.6 | 85.5 | 32.8 | 19.3 |
| Pose + OF Augmented RGB | 24.9 | 86.3 | 33.7 | 23.8 |

Poses, the RGB stream now dubbed as **Pose + OF Augmented RGB** learns some additional information (+2.6%, +6.7%, +4.5%, +4.9% w.r.t. vanilla RGB stream on Charades, PKU-MMD, TSU-CS, TSU-CV respectively).

**Inference time & Complexity**: Fig. 3 shows the precision vs inference time per video on Charades dataset. The inference time includes the time of extracting the additional modalities and the processing time of the visual encoder and temporal filter. We find that Two stream RGB + TVL1 [64] achieves high precision on Charades, but at the expense of a high computational cost. In this work, we use TVL1 to obtain the OF modality. Although there are methods [27, 52] that generate OF at higher speed, these methods perform significantly worse than TVL1 [6]. Similarly, the computation of accurate 3D Poses is not real-time, and hence doubles the video processing time [44]. With these modalities in training phase, our proposed framework avoids estimating these modalities at test time while keeping the performance of two-stream network. The processing speed at the inference phase (I3D+SSTCN) is about 140 fps using 4 GPUs, thus can be seen as a real-time processing. Concerning complexity, as we have the same type of temporal filter and encoder for teacher and student, the Augmented-RGB stream retains the same number of parameters as the vanilla RGB stream at inference time, whereas two stream network doubles the number of parameters often causing over-fitting [59].

### 4.5. Qualitative Analysis

With Global Contextual Relation loss, the student learns the relationships among the action instances of the teacher network along with retaining the student's individual representation. As shown in Fig. 4, with only $\mathcal{L}_{Global}$, the channel covariance representation of Augmented-RGB is closer to the one of RGB+OF. Hence, the Augmented-RGB achieves performance close to the one of the two-stream network.

We also compare the performance of RGB stream with Boundary Saliency distillation and vanilla RGB stream. In Fig. 5, we find that the network with $\mathcal{L}_{Boundary}$ detects tighter temporal boundaries of the actions compared to the vanilla network. To further show how two sequence-level distillation losses are complementary, we compare APs for a student that is trained with only $\mathcal{L}_{Global}$ or $\mathcal{L}_{Boundary}$ on Charades in Fig. 6. We find that $\mathcal{L}_{Boundary}$ improves more the actions with high variation across time (e.g. *Throw pillow*), $\mathcal{L}_{Global}$ improves more the actions with relatively longer duration (e.g. *Holding mirror*). While learning from $\mathcal{L}_{Global}+\mathcal{L}_{Boundary}$, the student improves all action types, reflecting how these two loss-terms complement each other.

Fig. 7 shows the class-wise actionness result of the vanilla-RGB and Augmented-RGB in a densely labeled video along with the action detection results. We notice that the Augmented-RGB detects tight action boundary w.r.t. the vanilla-RGB, e.g. *use cupboard, walk*. Thanks to our distillation methods, the Augmented-RGB now predicts the *use drawer* action which is miss detected in vanilla-RGB.

### 4.6. Comparison with the State-of-the-Art

In Tab. 6, we compare other action detection methods with our Augmented-RGB on PKU-MMD. Recall that our distillation mechanisms are build on SSTCN. While one method [32] using Poses achieves very high performance, this method is skeleton-based and applicable only for specific datasets (i.e. NTU-RGBD [45], PKU-MMD [36]), where high quality 3D Poses are available. In contrast, our method is generic and does not rely on Poses at inference time while being more effective compared to other RGB based SoA methods, such as Graph Distillation [38] (+2.6%, +2.4%, +4.6% for 0.1, 0.3, 0.5 IoU), which utilizes the same temporal filter but more modalities (e.g. depth) at

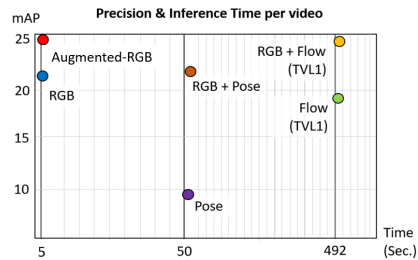

Figure 3. Precision vs Inference time per video on Charades.

Table 6. Event-based mAP on PKU-MMD (CS) dataset. Only the last five rows utilize RGB at inference time. Note that Graph distillation (GD) learns from more than 4 modalities while our method learns from OF and Pose.

| | | | mAP@tIoU ($\theta$) | | |
|---|---|---|---|---|---|
| | | Method | 0.1 | 0.3 | 0.5 |
| Test modality | Poses | JCRRNN [33] | 45.2 | – | 32.5 |
| | | Convolution Skeleton [36] | 49.3 | 31.8 | 12.1 |
| | | Skeleton boxes [1] | 61.3 | – | 54.8 |
| | | Wang and Wang [57] | 84.2 | – | – |
| | | Li et al. [32] | 92.2 | – | 90.4 |
| | RGB | Deep RGB [36] | 50.7 | 32.3 | 14.7 |
| | | Qin and Shelton [43] | 65.0 | 51.0 | 29.4 |
| | | GRU+GD [38] | 82.4 | 81.3 | 74.3 |
| | | SSTCN+GD | 83.7 | 82.1 | 76.5 |
| | | Augmented-RGB | **86.3** | **84.5** | **81.1** |

Table 7. Comparison with State-of-the-Art action detection methods. Our method learns only from OF. The cells in white are the two stream results (RGB+OF), while the cell in orange represents using only RGB at Inference time. We report frame-based mAP and event-based mAP for the dense and sparse labeled datasets respectively. The IoU is 0.5 for THUMOS14.

| Type | Model | Dense | | | Sparse |
|---|---|---|---|---|---|
| | | Charades | TSU-CS | MultiTHUMOS | THUMOS14 |
| Anchor | R-C3D [61] | 12.7 | 8.7 | — | 28.9 |
| | TAL [3] | — | — | — | 42.8 |
| | G-TAD [62] | — | — | — | 40.2 |
| | AFNet [5] | 13.1 | — | — | 49.5 |
| Seq2Seq | TAN [10] | 17.6 | — | 33.3 | 46.8 |
| | WSGM [15] | 18.7 | — | — | 32.8 |
| | TGM [42] | 21.5 | 26.7 | 44.3 | 53.5 |
| | Vanila-RGB [14] | 22.3 | 29.2 | 37.8 | 46.1 |
| | Two-stream | 24.4 | **33.5** | 44.4 | **53.7** |
| | Augmented-RGB | **24.6** | 32.8 | **44.6** | 53.3 |

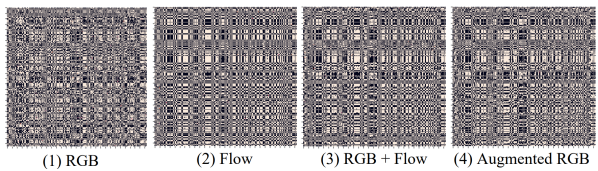

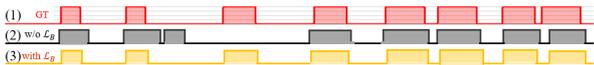

(1) RGB (2) Flow (3) RGB + Flow (4) Augmented RGB

Figure 4. Channel Covariance. We visualize the Covariance matrix of a video for the vanilla RGB, vanilla OF, the two-stream RGB+OF, and the Augmented-RGB ($\mathcal{L}_{Global}$). For better visualization, we normalize the matrix to [0,1] and set a threshold of 0.5.

(1) GT
(2) w/o $\mathcal{L}_B$
(3) with $\mathcal{L}_B$

Figure 5. Action boundary detection: (1) Ground truth indicates if it is action or background at this frame. (2) The boundaries detected without $\mathcal{L}_{Boundary}$, (3) The boundaries detected with $\mathcal{L}_{Boundary}$.

training time compared to our method.

To show the generalization of our method, we also evaluate our distillation framework on Charades and TSU-CS, MultiTHUMOS and THUMOS in Table 7. For all these comparisons, the student network is distilled with teacher pre-trained with OF in the training phase, as Poses are not always available. For a fair comparison with our Augmented-RGB, Vanilla-RGB and Two-stream networks are implemented using SSTCN. In this table, we find that, anchor-based methods (e.g. AFNet) perform decently on sparsely-labeled datasets, while failing on densely labeled datasets due to the combinatorial explosion of proposals. On the other hand, Seq2Seq architectures are stable on both types of dataset. With the help of our proposed distillation method, the Augmented-RGB achieves the competitive Two-stream performance on all the datasets (+2.3, 3.6, 6.8, 7.2 w.r.t. vanilla-RGB on Charades, TSU, Multi-THUMOS, THUMOS14 respectively). We observe that the performance improvement on THUMOS which consists of sport videos, is significant due to strong motion patterns resulting to an effective OF based teacher network. Thus, Augmented-RGB while using only RGB at inference, performs on par with Two-stream network for the task of action detection.

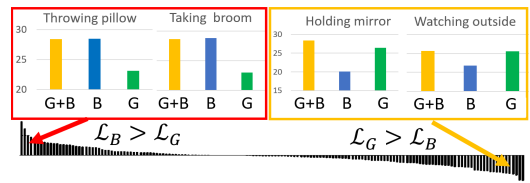

Figure 6. Difference of Average Precision for two sequence-level distillation losses on Charades dataset. G: $\mathcal{L}_{Global}$, B: $\mathcal{L}_{Boundary}$.

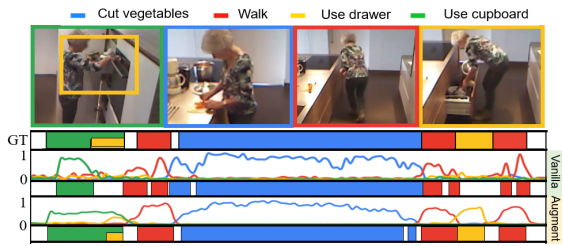

Figure 7. Class-wise actionness with the detection results.

## 5. Conclusion

In this work, we have introduced a novel distillation framework for action detection. This generic framework combines three novel learnable losses to better benefit from the cross-modal information in untrimmed videos. To the best of our knowledge, we are the first to propose a formulation with sequence-level distillation in this task. Thanks to this framework, we can improve the performance of vanilla RGB networks and make it possible to detect actions in real-time with high precision, even in case of densely labeled datasets. Experiments on five datasets show that the proposed method can effectively infuse different modalities into RGB. For instance, the Augmented-RGB network achieves the Two-stream network performance while using only RGB at inference time.

**Acknowledgements.** This work has been supported by the French government, through the 3IA Côte d'Azur Investments in the Future project managed by the National Research Agency (ANR) with the reference number ANR-19-P3IA-0002. The authors are also grateful to the OPAL infrastructure from Université Côte d'Azur for providing resources and support.